\documentclass[conference]{IEEEtran}
\IEEEoverridecommandlockouts

\usepackage{cite}
\usepackage{amsmath,amssymb,amsfonts}
\usepackage{algorithmic}
\usepackage{graphicx}
\usepackage{textcomp}
\usepackage{xcolor}
\usepackage{xurl}
\usepackage{booktabs}
\usepackage{array}
\usepackage{microtype}
\usepackage{listings}
\usepackage{orcidlink}
\usepackage{fancyhdr}

\graphicspath{{figures/}}

\fancypagestyle{arxivnotice}{%
  \fancyhf{}%
  \fancyfoot[C]{\parbox{\textwidth}{\centering\footnotesize
    This work has been accepted for publication in AIxB 2026. Copyright may be
    transferred without notice, after which this version may no longer be accessible.}}%
}

\lstdefinestyle{gtecode}{
  basicstyle=\ttfamily\footnotesize,
  breaklines=true,
  breakatwhitespace=false,
  columns=fullflexible,
  keepspaces=true,
  frame=none,
  showstringspaces=false,
  language=Python
}

\def\BibTeX{{\rm B\kern-.05em{\sc i\kern-.025em b}\kern-.08em
    T\kern-.1667em\lower.7ex\hbox{E}\kern-.125emX}}

\begin{document}

\title{Governing the Edge: Automating Commercial\\Property and Casualty Insurance Underwriting\\via a Hybrid Local-Cloud Multi-Agent Framework}

% \IEEEpubid{\makebox[\columnwidth]{978-1-XXXX-XXXX-X/26/\$31.00~\copyright2026 IEEE \hfill}
%   \hspace{\columnsep}\makebox[\columnwidth]{}}

\author{
\IEEEauthorblockN{Vivek Kumar Singh}
\IEEEauthorblockA{\textit{Independent Researcher}\\
McKinney, TX, USA\\
\orcidlink{0009-0002-9350-3207}\,0009-0002-9350-3207\\
vivekksingh.nov12@gmail.com}
\and
\IEEEauthorblockN{Gautam Bhowmick}
\IEEEauthorblockA{\textit{Independent Researcher}\\
Chicago, IL, USA\\
\orcidlink{0009-0001-9424-7826}\,0009-0001-9424-7826\\
bhowmickgautam@gmail.com}
}

\maketitle
\thispagestyle{arxivnotice}

%==============================================================================
\begin{abstract}
Underwriters in commercial Property and Casualty (P\&C) insurance spend most of
their day on paperwork rather than on judgment. Industry studies estimate that 30
to 40\% of an underwriter's time goes into administrative work such as reading
documents, looking up data, and rekeying, and a single submission can take
about 40 minutes to process by hand. This paper presents Governing the Edge, a
multi-agent framework that takes over that administrative layer. It uses 13 nodes (11 specialized agents and 2 deterministic control nodes, one of them
a human-escalation interrupt) arranged as a directed workflow graph in
LangGraph, and it splits the work across two tiers. Agents that touch sensitive
data run locally on Gemma 2 and are each bound to an authoritative source ---
flood maps, motor vehicle records, business-credit profiles, and federal
motor-carrier safety records --- through a tool-call interface that records every
invocation in an audit log (the external tools are synthetic stubs in the current
prototype). Agents that only
need to reason over anonymized scores run on Claude Sonnet through LiteLLM. The
part we think matters most is that compliance is built into the structure of the
graph. Rules act as conditions on the edges between agents, so a non-compliant
submission cannot reach pricing in the first place, instead of being caught
after the fact. We walk through one full scenario end to end, a commercial auto
submission whose principal driver carries serious violations, showing every agent
call, every tool invocation, and the routing decision that results. In the scenarios we ran,
the framework processes a submission that needs about 40 minutes of manual work in
a few minutes on a single edge device, with serialized on-device Gemma 2
inference dominating that time and a production graphics processing unit (GPU)
expected to reduce it. The
accuracy comes at the cost of latency: on-device inference makes the framework
slower per submission than a single cloud prompt, a trade we quantify and discuss.
On the hard-stop violation tier --- the cases a regulator most cares about ---
the framework enforces every rule correctly and reproducibly, because hard stops
are deterministic predicates over fields extracted at temperature~0; its overall
compliance accuracy (percentage of scenarios decided correctly)
across the 20-scenario benchmark is 70\%, with the remaining
gap concentrated in softer, judgment-based tiers. It keeps a complete audit trail
and respects the privacy boundary throughout, and includes per-agent error-tracking
infrastructure across eight failure categories for operational monitoring. We release all of the code, the compliance
rules, the tool stubs, and the synthetic datasets.
\end{abstract}

\renewcommand{\IEEEkeywordsname}{Keywords}

%==============================================================================
\section{Introduction}
Underwriting a commercial P\&C policy takes time and carries real consequences if
it goes wrong, and most of that time is not spent on judgment. McKinsey estimates
that in many commercial books, 30 to 40\% of an
underwriter's time goes into administrative work such as rekeying data and reading
documents rather than evaluating risk~\cite{iir}, so the bulk of the elapsed time
sits in handling the submission rather than in the risk decision itself.

This tells us where automation should aim: not at the underwriter's judgment but
at the administrative work in front of it. A typical 40-minute submission splits
into about 10 minutes reading documents and pulling fields, 15 on data
lookups against the motor vehicle record (MVR), the Federal Emergency Management
Agency (FEMA) flood service, the Department of Transportation (DOT), and
Dun \& Bradstreet (D\&B), 10 checking compliance rules, and 5
writing up the audit record. The risk call itself, the thing underwriters train
for, is a small slice of the total.

Existing AI approaches only cover part of this. A single large prompt can read a
submission and produce a quote, but it cannot enforce compliance, cannot call out
to authoritative data, and usually ships raw sensitive data to a cloud application
programming interface (API).
Agent pipelines such as those built with CrewAI~\cite{sajid} do split the work
into separate agents, but they coordinate on task dependencies rather than on
compliance, and the agents reason over the submission text instead of querying
authoritative data sources. Neither keeps data private, leaves a regulator-grade
audit trail, nor grounds decisions in verified outside data.

Governing the Edge is our attempt to fix those gaps, and it rests on four design
choices. We write the compliance rules directly into the graph as conditions on
the transitions between agents. We split the system across a local and a cloud
tier, keeping the privacy-sensitive agents on-premise with Gemma 2. We give each
agent a dedicated tool boundary to an external source --- stubbed in this prototype
--- and log every call. And we record where each agent fails, sorted into eight
categories, so we can monitor reliability over time rather than infer it indirectly.

Our contributions are the following. We design a governance-aware multi-agent
underwriting architecture for commercial P\&C on top of LangGraph, and we pair it
with a local-cloud split that keeps sensitive data on-premise. We build a 10-rule
P\&C compliance engine that covers both commercial property and commercial auto,
checked against real underwriting practice, with hard-stop rules enforced by
deterministic predicates, along with a tool-call layer that binds each agent to an
authoritative production data source (synthetic stubs in the released prototype). We add per-agent error-tracking
infrastructure spanning eight failure categories for operational monitoring. We
work through a complete scenario showing the system automate a piece of
underwriting end to end, turning roughly 40 minutes of manual work into a few
minutes of automated execution. Submissions that clear every check undergo
straight-through processing --- they pass from intake to final decision with no
human touch --- while the rest are routed to an underwriter. Finally, we release the prototype and a synthetic
benchmark as open source.

Rules were developed with guidance from a technology practice leader in insurance
application platforms. Production deployment requires updating thresholds to
reflect state-specific Department of Insurance (DOI) requirements and National
Association of Insurance Commissioners (NAIC) model laws.

%==============================================================================
\section{Related Work}
\label{sec:related}
\subsection{AI in Insurance Underwriting}
Sajid~\cite{sajid} builds a CrewAI system for evaluating property claims
and reports up to 92.9\% accuracy in claim evaluation. The agents are coordinated
over a task graph rather than a single line, but routing is driven by task
dependencies, not compliance, and the agents reason over document text without
authoritative external tools; our Linear Pipeline baseline
(Section~\ref{sec:baselines}) is a sequential, compliance-at-the-end reduction in
this spirit. Roy and Singh~\cite{bharadwaj} take a different
angle, pairing a primary agent with an adversarial critic in a human-in-the-loop
setup, using a state machine and retrieval-augmented generation (RAG) over an
underwriting manual, and contribute a failure-mode taxonomy for the
decision-making agent.
% "decision-making agent" as the most likely intended reading. Confirm the exact
% term against arXiv:2602.13213 and correct if it denotes something specific.
% "failure-mode taxonomy" against the body of arXiv:2602.13213.
Their goal of reliability in regulated underwriting is close to ours,
but their system runs entirely in the cloud, keeps a human in the binding loop
rather than enforcing compliance structurally in the graph, and does not address
the privacy boundary or grounding in external authoritative data that we focus on.

\subsection{Multi-Agent Large Language Model (LLM) Frameworks}
LangGraph~\cite{langgraph} gives us graph-based orchestration with shared state,
conditional edges, and interrupts, and those are exactly the features that let us
express governance as routing. AutoGen~\cite{autogen} coordinates agents through
conversation, which brings a degree of randomness that we did not want in a
regulated decision. Other multi-agent designs target coordination efficiency and
architectural flexibility through task-distribution and collaboration patterns,
but none route on compliance or enforce a privacy
boundary. RouteLLM~\cite{routellm} and FrugalGPT~\cite{frugalgpt} are
about routing a single call to the right model, not about governing a multi-step
workflow. Like these systems we route, but at the level of the whole workflow
rather than a single call: we decide the governance path at plan time, before
execution, rather than reacting node by node.

\subsection{Privacy-Preserving and Edge AI}
Xu et al.~\cite{xu} survey on-device language models and the architectural and
compression techniques that make running capable LLMs locally practical, motivated
in part by data localization and privacy, exactly the constraints that arise when
sensitive submission data cannot leave the perimeter. Gemma 2 runs locally and is
small enough to deploy on-premise~\cite{gemma}, which makes it a good fit for the
sensitive parts of our pipeline. We carry the same edge-AI idea into insurance and
draw an explicit line between what runs locally and what runs in the cloud.

%==============================================================================
\section{System Architecture}
\label{sec:arch}
\subsection{Overview}
\begin{figure}[tbp]
\centering
\includegraphics[width=\columnwidth]{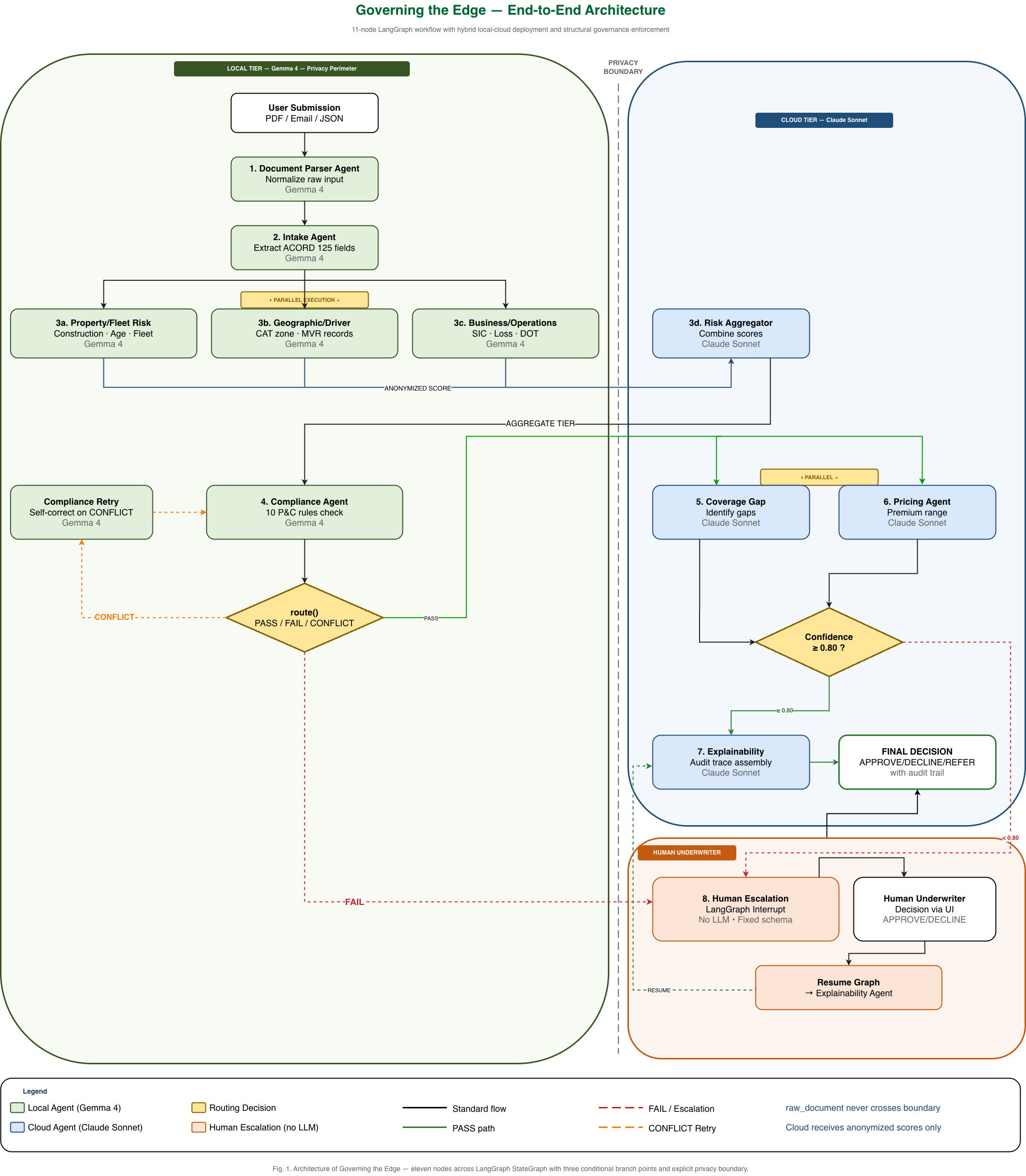}
\caption{End-to-end architecture of the Governing the Edge framework. Thirteen
nodes execute across a LangGraph StateGraph with three conditional branch points
and an explicit privacy boundary separating raw submission data (local tier) from
anonymized score processing (cloud tier).}
\label{fig:arch}
\end{figure}

The framework is a LangGraph StateGraph with 13 nodes and three points where
the path can branch, shown in Fig.~\ref{fig:arch}. The agents sit in two tiers.
The local tier runs Gemma 2 (\texttt{ollama/gemma2}) on-premise and handles
anything that touches sensitive data, while the cloud tier runs Claude Sonnet~4.6
(\texttt{claude-sonnet-4-6}) through LiteLLM and handles the heavier reasoning over
data that has already been anonymized.
% If you re-run on Gemma 4, update this and the abstract/intro/Related Work, and
% regenerate evaluation/results.json so the latency numbers reflect Gemma 4.

The flow runs roughly top to bottom. The Document Parser (1) cleans up the raw
PDF or email, and the Intake Agent (2) pulls out the fields of the ACORD~125
form, the standardized commercial insurance application defined by the Association
for Cooperative Operations Research and Development (ACORD) standards body. Three risk
sub-agents (3a, 3b, 3c) then run in parallel, each invoking the tool bound to its
dimension. The Risk Aggregator (3d) is the first cloud agent and merges their scores.
The Compliance Agent (4) checks the rules engine. If that returns PASS, Coverage
Gap (5) and Pricing (6) run in parallel, their outputs are combined at a
parallel-merge node, and the Explainability Agent (7) puts
together the audit trace. The Human Escalation node (8) is a plain interrupt,
with no model call, that fires whenever the confidence score drops below 0.80 or
compliance fails. Table~\ref{tab:agents} lists these 11 agent nodes and the
human-escalation interrupt; two further deterministic control nodes complete the
StateGraph --- a compliance-retry node (Section~\ref{sec:gov}) and the parallel-merge
step just mentioned --- bringing the total to 13.

\begin{figure}[tbp]
\centering
\includegraphics[width=\columnwidth]{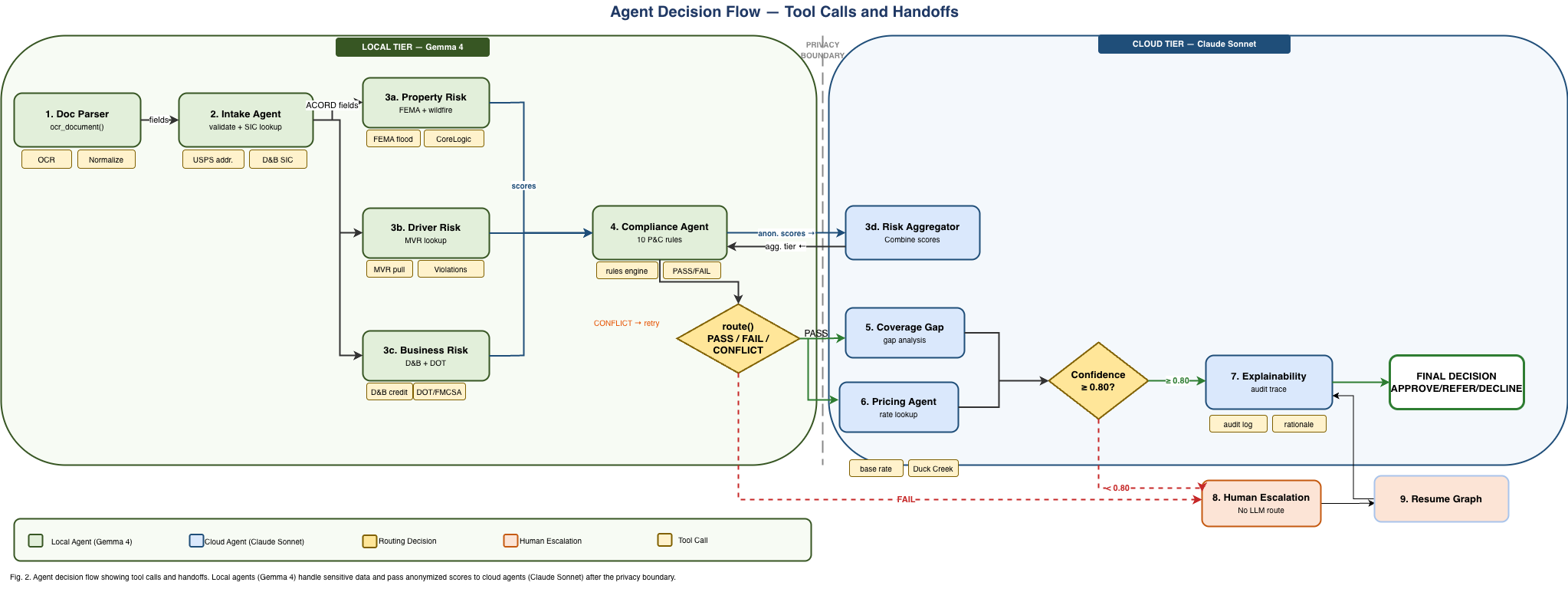}
\caption{Agent decision flow. Local agents (Gemma 2) invoke their bound tool
interfaces and pass anonymized scores to cloud agents (Claude Sonnet) after the
privacy boundary.}
\label{fig:decision}
\end{figure}

Fig.~\ref{fig:decision} shows the agent decision flow with tool calls and
handoffs. Local agents invoke their bound tool interfaces and pass structured
scores to the Risk Aggregator, the first Claude Sonnet agent, which crosses the
privacy boundary receiving only anonymized scores. The Compliance Agent then runs
the 10 P\&C rules and routes the workflow to PASS, FAIL, or CONFLICT, with PASS
triggering parallel Coverage Gap and Pricing agents on the cloud tier before a
confidence check (threshold 0.80 on a 0--1 confidence scale) decides between a
Final Decision and Human
Escalation.

\begin{table}[tbp]
\caption{Agent Inventory (two control nodes omitted; total = 13).}
\label{tab:agents}
\centering
\scriptsize
\setlength{\tabcolsep}{3pt}
\renewcommand{\arraystretch}{1.0}
\begin{tabular}{@{}clllp{2.4cm}@{}}
\toprule
\textbf{\#} & \textbf{Agent} & \textbf{Model} & \textbf{Tier} & \textbf{Role}\\
\midrule
1   & Doc Parser      & Gemma 2       & Local & Normalize input\\
2   & Intake          & Gemma 2       & Local & Extract ACORD 125 (temp=0)\\
3a  & Property/Fleet  & Gemma 2       & Local & FEMA/wildfire stubs\\
3b  & Geo/Driver      & Gemma 2       & Local & MVR stub; driver risk\\
3c  & Business/Ops    & Gemma 2       & Local & D\&B/DOT stubs\\
3d  & Risk Aggregator & Claude Sonnet & Cloud & Merge scores (anon.)\\
4   & Compliance      & Py+Gemma 2    & Local & Hard-stop preds.; WARNs\\
5   & Coverage Gap    & Claude Sonnet & Cloud & Coverage gaps\\
6   & Pricing         & Claude Sonnet & Cloud & Premium range\\
7   & Explainability  & Claude Sonnet & Cloud & Audit trace\\
8   & Escalation      & None          & Local & Deterministic interrupt\\
\bottomrule
\end{tabular}
\end{table}

\subsection{Privacy Boundary}
The privacy line is drawn where state is serialized. The sensitive fields,
\texttt{raw\_document} and \texttt{parsed\_fields}, are omitted from the state
object sent to the cloud agents, which see only the anonymized
\texttt{risk\_profile}, \texttt{compliance\_result}, and \texttt{coverage\_gaps}.
We enforce this in three independent places: the serializer that drops the fields,
the cloud prompt wording, and network-level isolation. Tool calls follow the same
rule: MVR pulls, credit checks, and anything sensitive run only locally, and cloud
agents never call a tool needing sensitive input.

\subsection{Provider-Agnostic LLM Layer}
Every model call goes through LiteLLM\footnote{\url{https://github.com/BerriAI/litellm}},
so the rest of the code does not care which
provider is behind it, and the model can be set per agent (local agents default to
\texttt{ollama/gemma2}, cloud agents to \texttt{claude-sonnet-4-6}). Each agent's
output is checked against a Pydantic schema before it is allowed into shared state,
which means a malformed response is caught at the boundary rather than downstream.

%==============================================================================
\section{Governance Framework}
\label{sec:gov}
\subsection{Structural vs. Post-Hoc Governance}
Our central argument is that encoding compliance rules in the graph's transition
conditions catches more violations than checking after the fact. In a linear
pipeline a bad intermediate result simply propagates through risk assessment,
pricing, and coverage analysis, unnoticed until the end.

In our setup the edge from the Compliance Agent to Coverage Gap and Pricing only
opens when the check comes back PASS.

\begin{lstlisting}[style=gtecode]
def route_compliance(state) -> str:
    if state.compliance_result.status == 'PASS':
        return 'coverage_gap'      # proceed
    elif state.compliance_result.status == 'FAIL':
        return 'human_escalation'  # block
    else:  # CONFLICT
        return 'compliance_retry'  # self-correct
\end{lstlisting}

So the downstream agents are simply unreachable from a non-compliant state, not
skipped by a check but never routed to at all. A CONFLICT result routes to a
dedicated compliance-retry node that re-runs the check with a fuller prompt; if
the retry still does not return PASS, the workflow escalates to a human. We cap
this at one automated retry (\texttt{MAX\_COMPLIANCE\_RETRIES = 1}) before escalation.

The hard-stop rules are enforced by deterministic Python predicates
(\texttt{evaluate\_hard\_stops} in \texttt{governance/deterministic\_rules.py}),
not by model inference. The Compliance Agent runs these predicates first and only
routes the soft WARNING rules to Gemma~2, so a hard-stop decision never depends on
an LLM's output; the only path to a wrong hard-stop status is an upstream
field-extraction error, which makes the language model the error source rather than
the adjudicator. This is a stronger guarantee than an LLM-judged compliance step,
and, because the fields those predicates read are extracted at temperature~0, it is why hard-stop outcomes are stable across runs.

\subsection{P\&C Compliance Rules Engine}
The rules engine holds 10 P\&C underwriting rules covering both commercial
property and commercial auto, written and sanity-checked against real
underwriting practice. Table~\ref{tab:rules} lists them. Total insured value
(TIV) in rule CP-02 is the combined replacement cost of buildings, contents, and
business income at a single location.

\begin{table}[tbp]
\caption{P\&C Compliance Rules Engine: 10 Validated Underwriting Rules}
\label{tab:rules}
\centering
\scriptsize
\renewcommand{\arraystretch}{1.0}
\begin{tabular}{@{}llp{3cm}ll@{}}
\toprule
\textbf{ID} & \textbf{Line} & \textbf{Rule} & \textbf{Sev.} & \textbf{Agent}\\
\midrule
CP-01 & Prop. & Frame const.\ $>$3 stories         & STOP & 3a\\
CP-02 & Prop. & TIV $>$ USD~10M                       & STOP & 3a\\
CP-03 & Prop. & Habitational $>$4 units             & STOP & 3a\\
CP-04 & Prop. & Loss ratio $>$70\%                  & WARN & 3a\\
CP-05 & Prop. & Bldg $>$40yr, no reno              & WARN & 3a\\
CA-01 & Auto  & Fleet $>$20 vehicles                & STOP & 3a\\
CA-02 & Auto  & Driver 2+ major viol.\ (3yr)        & STOP & 3b\\
CA-03 & Auto  & Radius $>$500 mi                    & WARN & 3b\\
CA-04 & Auto  & Avg fleet $>$10yr                   & WARN & 3a\\
CA-05 & Auto  & DOT Cond./Unsat.                    & STOP & 3c\\
\bottomrule
\end{tabular}
\end{table}

A HARD STOP rule sends the submission straight to a human no matter how confident
the system is. A WARNING rule does not stop anything on its own, but it lowers the
confidence score and gets written into the audit trail.

%==============================================================================
\section{Tool Integration}
\label{sec:tools}
A recurring weakness in LLM agent systems is that agents reason only from the text
in front of them. Our architecture gives each agent a dedicated tool boundary to an
authoritative source, and logs every call as a structured ToolCall record. In the
current prototype these tools are synthetic stubs (Section~\ref{sec:limits}); the
contribution at this stage is the tool-call interface and audit plumbing, not live
data grounding. Table~\ref{tab:tools} maps each agent to the production API its
stub stands in for. We are explicit about a boundary that matters for evaluation:
the deterministic hard-stop rules read fields extracted from the submission
document (for example \texttt{driver\_mvr\_violations}), not the soft scores
returned by the risk tools, so in this prototype the hard-stop decisions are
grounded in document extraction rather than in the stubbed external lookups.
Wiring the rules to consume verified tool output instead of extracted fields is a
planned step toward true external grounding. Document parsing relies on optical
character recognition (OCR) to extract text from scanned forms before Gemma~2
normalizes it.

\begin{table*}[!t]
\caption{Tool Interface Map: Agent to Authoritative Production Source (synthetic stubs in prototype). Source acronyms: USPS, United States Postal Service; NAICS, North American Industry Classification System; FMCSA, Federal Motor Carrier Safety Administration.}
\label{tab:tools}
\centering
\footnotesize
\renewcommand{\arraystretch}{1.05}
\begin{tabular}{@{}lll@{}}
\toprule
\textbf{Agent} & \textbf{Tool Function} & \textbf{Production API}\\
\midrule
Document Parser Agent & \texttt{ocr\_document()} & AWS Textract / Azure Form Recognizer\\
Intake Agent & \texttt{validate\_address()}, \texttt{lookup\_sic\_code()} & USPS Address Validation; D\&B SIC/NAICS\\
Property Risk & \texttt{lookup\_fema\_flood\_zone()}, \texttt{lookup\_wildfire\_risk()} & FEMA Flood Map; CoreLogic Wildfire / Verisk\\
Geographic/Driver & \texttt{lookup\_mvr()} & LexisNexis MVR / Equifax (consent required)\\
Business Risk & \texttt{lookup\_dnb\_business()}, \texttt{lookup\_dot\_safety\_rating()} & D\&B Hoovers / Experian; FMCSA SAFER\\
Compliance Agent & \texttt{query\_rules\_engine()} & Internal rules DB + state DOI filings\\
Pricing Agent & \texttt{lookup\_base\_rate()} & Duck Creek Rating / Guidewire PolicyCenter\\
\bottomrule
\end{tabular}
\end{table*}

Each call is logged with timestamp, inputs, outputs, success flag, and duration.
The system maps failures to eight categories (Table~\ref{tab:errors}); on the released
20-scenario run all executions succeeded, so this is operational infrastructure rather
than a measured result.

\begin{table}[tbp]
\caption{Per-Agent Error Taxonomy: The Eight Failure Categories Tracked}
\label{tab:errors}
\centering
\footnotesize
\renewcommand{\arraystretch}{1.15}
\begin{tabular}{@{}lp{5.2cm}@{}}
\toprule
\textbf{Category} & \textbf{Trigger}\\
\midrule
\texttt{TIMEOUT}            & LLM call exceeded its timeout.\\
\texttt{LLM\_API\_ERROR}    & Authentication, rate-limit, or quota error from the model API.\\
\texttt{LLM\_PARSE\_ERROR}  & JSON/output parsing failed (also the catch-all default).\\
\texttt{SCHEMA\_VALIDATION} & Output failed Pydantic schema validation.\\
\texttt{TOOL\_FAILURE}      & Tool/function-call invocation failed.\\
\texttt{INCONSISTENCY}      & Output inconsistency or contradiction detected.\\
\texttt{NULL\_REQUIRED}     & A required field was null or missing.\\
\texttt{GOVERNANCE\_CONFLICT} & A governance/compliance rule was violated.\\
\bottomrule
\end{tabular}
\end{table}

%==============================================================================
\section{Experimental Walkthrough Scenario}
\label{sec:scenario}
To make this concrete, we follow one commercial auto submission from raw input to
the routing decision, with every agent call and every tool call.
The submission is scenario CA-S06, a three-vehicle courier service whose principal
driver has a string of serious violations; it touches every tool type and trips a
hard-stop rule (CA-02).

\subsection{Input Submission}
Scenario CA-S06: Speedy Couriers Inc (Standard Industrial Classification (SIC)
code 4215, Chicago IL), fleet of 3 cargo vans
(avg age 4 yr), radius 120 mi. Principal driver: driving under the influence (DUI)
2021, reckless driving 2022,
at-fault accident 2023 --- three major violations in three years.
Coverage: Liability USD~500K / Physical Damage USD~75K.

\textit{Steps 1--2.} Document Parser normalizes the PDF (ACORD\_125, high confidence).
Intake Agent calls \texttt{validate\_address()} and \texttt{lookup\_sic\_code()}, then
extracts fleet\_size = 3, avg\_fleet\_age = 4, radius = 120, and
driver\_mvr\_violations = 3 (worst driver, three major violations in three years).

\textit{Step 3 --- Parallel Risk Sub-Agents.} The three
sub-agents run concurrently. Property/Fleet Risk (3a) has no external tool on a
commercial-auto submission (its FEMA and wildfire tools are gated to commercial
property), so it scores on fleet makeup and age, returning a MEDIUM risk band (a
discrete risk label spanning LOW, MEDIUM, HIGH, and DECLINE) and citing
the cargo-van composition and four-year average age.

Geographic/Driver Risk (3b) is decisive here. It invokes the MVR tool stub and
scores the driver-risk dimension, returning a DECLINE band; the CA-02 hard-stop
itself is enforced downstream at the Compliance Agent, which reads the extracted
driver\_mvr\_violations = 3 field (two or more major violations in three years).

Business/Ops Risk (3c) calls \texttt{lookup\_\allowbreak dnb\_\allowbreak business()}
(years\_in\_business = 8, credit\_score = 78 on a 0--100 scale, no liens), then
\texttt{lookup\_\allowbreak dot\_\allowbreak safety\_\allowbreak rating()}, which
returns NOT\_RATED with success = False (no DOT number provided). It records the
missing input and settles on a MEDIUM band with reduced confidence.

\textit{Step 4 --- Risk Aggregator Crosses Privacy Boundary.}
The Risk Aggregator is the first cloud agent, and all it sees is the three
sub-agent scores and their band labels under the commercial-auto line weights
(fleet 0.30, driver 0.45, ops 0.25; weights summing to 1). No driver names, no business name, and no
violation details cross the boundary. The aggregator is a Claude Sonnet agent: it
returns an elevated risk tier driven by the dominant driver-risk band. Because the
hard-stop outcome in this scenario is determined deterministically at the
compliance step (Step~5), the exact aggregate value does not affect the decision.
% cloud_agents.py: risk_aggregator_node), NOT a deterministic weighted average,
% and the per-scenario aggregate is not logged in results.json. We therefore
% describe the tier qualitatively and let the deterministic CA-02 hard-stop carry
% the DECLINE, which is what the code actually does.

\textit{Step 5 --- Compliance Agent Triggers Hard-Stop.} The
Compliance Agent evaluates the commercial-auto rules for IL. The hard-stop checks
are deterministic Python predicates: \texttt{check\_ca02\_driver\_violations} reads
driver\_mvr\_violations = 3 and, because $3 \geq 2$, returns a CA-02 violation with
severity HARD\_STOP and action DECLINE. The compliance result is therefore FAIL,
with a CA-02 entry flagging the principal driver for exclusion. This decision does
not depend on any LLM output.

\textit{Step 6 --- Routing Decision.}
\texttt{route\_compliance()} sees status = FAIL and routes to Human Escalation;
Coverage Gap and Pricing never run. This is exactly where the design parts ways
with a linear pipeline, which would have priced a submission it should not have.

\textit{Step 7 --- Human Escalation Package.} The Human Escalation node (no model
call) assembles a fixed-schema package: submission\_id, escalation\_reason
(CA-02 detected), risk profile, compliance\_result with rule\_id, agent explanations,
audit trail, and recommended\_action DECLINE. Tool invocations are recorded in a
separate log so the underwriter can trace both model decisions and data evidence.

\textit{End-to-End Timing.} CA-S06 completed in 197.5 s (3.3 min); the 20-scenario
average is 268.8 s, dominated by serialized on-premise Gemma~2 inference
(Table~\ref{tab:automation}). A production edge GPU with resident model would
substantially reduce this. The same submission by hand takes 35--45 minutes.

\begin{table*}[!t]
\caption{Per-Agent Latency on the Released 20-Scenario Run (M1, serialized on-premise inference). Manual times are from the McKinsey commercial-underwriting benchmark~\cite{iir}; framework times are median wall-clock per agent from the shipped telemetry (\texttt{evaluation/error\_report.json}).}
\label{tab:automation}
\centering
\footnotesize
\renewcommand{\arraystretch}{1.05}
\begin{tabular}{@{}p{4.6cm}llp{5.8cm}@{}}
\toprule
\textbf{Underwriting Task} & \textbf{Manual Time} & \textbf{Framework (median)} & \textbf{Tool / Agent}\\
\midrule
Document intake \& field extraction & 16--24 min & 63.9 s & Doc Parser + Intake\\
Driver/geographic risk (incl.\ MVR) & 8--12 min & 71.7 s & Geographic/Driver (\texttt{lookup\_mvr()})\\
Property/fleet exposure check & 5--10 min & 80.3 s & Property/Fleet (\texttt{lookup\_fema\_flood\_zone()})\\
Business credit / DOT lookup & 10--18 min & 110.3 s & Business/Ops (\texttt{lookup\_dnb\_business()}, \texttt{lookup\_dot\_safety\_rating()})\\
Compliance rules check (10 rules) & 15--25 min & 39.2 s & Compliance (\texttt{query\_rules\_engine()})\\
Coverage adequacy analysis & 10--15 min & 46.8 s & Coverage Gap (cloud)\\
\midrule
\textbf{END-TO-END (avg over 20 scenarios)} & \textbf{$\sim$40 min} & \textbf{268.8 s} & Full Workflow\\
\bottomrule
\end{tabular}
\end{table*}

By automating the administrative layer that consumes 30 to 40\% of the
underwriter's day~\cite{iir}, the framework frees time for risk evaluation; on a
hard-stop case like CA-S06 it returns a complete, audit-ready package naming the
broken rule, the agent that caught it, and the recommended action.

%==============================================================================
\section{Comparison with Baseline Systems}
\label{sec:baselines}
We compare against two baselines. Monolithic Claude Sonnet is a single prompt
over the whole submission with no intermediate agents, tool calls, or compliance
step. The Linear Pipeline runs the same agents in sequence with compliance checked
only at the end and no tools. Both run entirely on cloud APIs.

On CA-S06, the monolithic baseline missed the CA-02 violation entirely, quoting a
USD~5{,}200 premium and never flagging the DUI because the violation count sat in
free-text history rather than a field the model attended to. The linear pipeline
caught it, but only at its final compliance check, after already pricing and
analyzing coverage for a submission it should have stopped.

Table~\ref{tab:baseline} reports per-system compliance accuracy, workflow
completion, and average latency across the 20 scenarios.

\begin{table}[tbp]
\caption{Baseline Comparison on 20 Synthetic Scenarios}
\label{tab:baseline}
\centering
\footnotesize
\renewcommand{\arraystretch}{1.05}
\begin{tabular}{@{}lccc@{}}
\toprule
\textbf{System} & \textbf{Compliance Acc.} & \textbf{Completion} & \textbf{Avg Latency}\\
\midrule
Monolithic Sonnet & 75.0\% & 100\% & 6.4 s\\
Linear Pipeline & 30.0\% & 100\% & 30.4 s\\
\textbf{Governing the Edge} & \textbf{70.0\%} & 100\% & 268.8 s\\
\bottomrule
\end{tabular}
\end{table}
% system. The numbers above match it exactly (0.75/0.30/0.70 acc;
% 6.4/30.4/268.8 s), with intake at temperature=0 so hard stops are reproducible.
% If you later run multiple seeds, report mean+/-range; otherwise keep single-run wording.

Compliance accuracy is the fraction of scenarios reaching the correct approve,
refer, or decline decision with all applicable hard-stop rules enforced; a
scenario routed to a human on an unresolved CONFLICT is scored
correct only when escalation is the appropriate outcome. Governing
the Edge reaches 70\% overall compliance accuracy, between the monolithic baseline
(75\%) and the linear pipeline (30\%). This single overall figure, however, understates
where the framework matters most. Table~\ref{tab:percomplexity} breaks accuracy
down by scenario tier. On the hard-stop violation tier --- the cases where a
non-compliant submission must be stopped --- Governing the Edge is correct on all
six scenarios (100\%), versus five of six (83\%) for the monolithic baseline; and
unlike the linear pipeline, which also catches all six hard stops but collapses to
0\% on every other tier, it does so without sacrificing the straightforward
approvals (4/4). The overall gap to the monolithic baseline is concentrated in the
WARNING and low-confidence escalation tiers (2/6 and 2/4), which are decided by the
local Gemma~2 model's judgment rather than by deterministic predicates; the
cloud-based monolithic Claude baseline is stronger on exactly those softer,
judgment-heavy cases. With 20 scenarios these tier-level differences span a handful
of scenarios each and are indicative rather than statistically established. The
hard-stop result, by contrast, is reproducible: it is computed by deterministic
predicates over temperature~0 fields and does not vary between runs.

\begin{table}[tbp]
\caption{Compliance Accuracy by Scenario Tier (committed 20-scenario run). Governing the Edge is the only system that enforces every hard stop (6/6) without collapsing on the other tiers.}
\label{tab:percomplexity}
\centering
\footnotesize
\renewcommand{\arraystretch}{1.05}
\begin{tabular}{@{}lccc@{}}
\toprule
\textbf{Scenario Tier} & \textbf{Monolithic} & \textbf{Linear} & \textbf{Gov. the Edge}\\
\midrule
Straightforward approval (4) & 4/4 & 0/4 & \textbf{4/4}\\
WARNING compliance (6) & 3/6 & 0/6 & 2/6\\
Hard-stop violation (6) & 5/6 & 6/6 & \textbf{6/6}\\
Low-confidence escalation (4) & 3/4 & 0/4 & 2/4\\
\midrule
\textbf{Overall} & \textbf{75\%} & \textbf{30\%} & \textbf{70\%}\\
\bottomrule
\end{tabular}
\end{table}

\subsection{Underwriter Handoff}
On escalation, underwriters receive a fixed-schema package with the exact rule, agent
evidence, and recommended action. The composite confidence score
(intra-agent 0.40, inter-agent 0.35, governance 0.25; weights summing to 1) routes uncertain
cases to a human at threshold 0.80 on the same 0--1 scale; the rest reach a Final Decision automatically.

% Ablation study removed for space; qualitative analysis available at the project repo.
%==============================================================================
\section{Limitations and Discussion}
\label{sec:limits}
\subsection{Limitations of the Current Prototype}
The prototype has several limitations that bear directly on how its results should
be read. First, the external tools are stubs that return synthetic, randomized data
rather than live lookups, and they are unseeded, so the cloud-tier soft scores, the
aggregate risk tier, and the per-run latencies vary between runs and are not
bit-for-bit reproducible. The hard-stop tier, computed by deterministic predicates
over temperature~0 fields rather than the randomized tool output, is stable (6/6
across runs) while the WARNING and escalation tiers vary; a fixed random number
generator (RNG) seed is a planned change that would make the entire run reproducible.
Second, because the
hard-stop rules read extracted fields rather than tool output, the tool layer
currently provides the call-and-audit interface but not yet live external
grounding for the decisions that matter; closing that gap is the main engineering
step toward production. Swapping in the real APIs would not touch agent prompts
or routing, since agents consume whatever structured dictionary the tool returns,
but that integration is until then untested.

On the model side, the released
configuration runs Gemma 2 locally; Gemma 4 is the intended target once on-device
tooling support lands, at which point the latency figures should be re-measured.
Risk aggregation is delegated to the cloud LLM rather than a fixed formula, so the
aggregate score is a model output rather than an auditable arithmetic step;
replacing it with a deterministic weighted sum is a planned simplification.

The scenarios are synthetic, which aids open release but limits
generalization. Ground-truth decisions were set by a single expert; a larger,
multiply-annotated benchmark would strengthen them. The manual times in
Table~\ref{tab:automation} come from a published benchmark~\cite{iir}
rather than our own carrier studies.

%==============================================================================
\section{Conclusion}
\label{sec:conclusion}
Governing the Edge encodes compliance as graph edge conditions with deterministic
hard-stop predicates, keeps sensitive agents on-premise on Gemma~2, and binds each
agent to a logged tool interface. On 20 synthetic scenarios it achieves 100\% hard-stop
accuracy (6/6, reproducibly) and 70\% overall, preserving full data residency and a
complete audit trail. Code, rules, stubs, and datasets:
\url{https://github.com/vsingh45/governing-the-edge}.

%==============================================================================
\section*{Acknowledgment}
The authors used a generative AI system (Anthropic Claude) during preparation of
this manuscript, limited to language editing and coding assistance. All technical
content, claims, experimental design, compliance rules, and architectural
decisions are entirely the authors' own work.

%==============================================================================
\enlargethispage{3\baselineskip}


\vspace{-4pt}
{\scriptsize
\begin{thebibliography}{00}
\setlength{\itemsep}{-2pt}
\setlength{\parsep}{0pt}
\setlength{\parskip}{0pt}
\setlength{\topsep}{0pt}
\bibitem{sajid} M. I. Sajid, ``Multi-agentic automation for evaluating property claims in underwriting,'' \textit{Open Journal of Applied Sciences}, vol. 15, no. 4, pp. 819--833, 2025, doi: 10.4236/ojapps.2025.154055.
\bibitem{bharadwaj} J. Roy and S. K. Singh, ``Agentic AI for commercial insurance underwriting with adversarial self-critique,'' arXiv preprint arXiv:2602.13213, 2026.
\bibitem{langgraph} LangChain, ``LangGraph: Building stateful, multi-actor applications with LLMs,'' 2024. [Online]. Available: \url{https://github.com/langchain-ai/langgraph}
\bibitem{autogen} Q. Wu et al., ``AutoGen: Enabling next-gen LLM applications via multi-agent conversation,'' arXiv preprint arXiv:2308.08155, 2023.
\bibitem{routellm} I. Ong et al., ``RouteLLM: Learning to route LLMs with preference data,'' arXiv preprint arXiv:2406.18665, 2024.
\bibitem{frugalgpt} L. Chen et al., ``FrugalGPT: How to use large language models while reducing cost and improving performance,'' arXiv preprint arXiv:2305.05176, 2023.
\bibitem{xu} J. Xu et al., ``On-device language models: A comprehensive review,'' arXiv preprint arXiv:2409.00088, 2024.
\bibitem{gemma} Gemma Team, Google DeepMind, ``Gemma 2: Improving open language models at a practical size,'' arXiv preprint arXiv:2408.00118, 2024.
\bibitem{iir} A. Agarwal, ``Fixing underwriting's front door,'' \textit{Insurance Innovation Reporter}, Jan. 14, 2026.
\end{thebibliography}
}
\end{document}